\documentclass{article}

\usepackage{microtype}
\usepackage{graphicx}
\usepackage{subfigure}
\usepackage{booktabs} 

\usepackage{amsfonts}
\usepackage{amsmath}
\usepackage{multirow}

\usepackage{hyperref}

\DeclareMathOperator*{\argmin}{argmin}
\newcommand{\xhdr}[1]{\paragraph*{\bf {#1.}}}

\newcommand{\Real}{\mathbb{R}}
\newcommand{\BigO}[1]{\ensuremath{\operatorname{O}\bigl(#1\bigr)}}

\usepackage[accepted]{icml2019}

\icmltitlerunning{Understanding the Origins of Bias in Word Embeddings}

\begin{document}

\twocolumn[
\icmltitle{Understanding the Origins of Bias in Word Embeddings}



\icmlsetsymbol{equal}{*}

\begin{icmlauthorlist}
\icmlauthor{Marc-Etienne Brunet}{to,vec}
\icmlauthor{Colleen Alkalay-Houlihan}{to}
\icmlauthor{Ashton Anderson}{to,vec}
\icmlauthor{Richard Zemel}{to,vec}
\end{icmlauthorlist}

\icmlaffiliation{to}{Department of Computer Science, University of Toronto, Toronto, Canada}
\icmlaffiliation{vec}{Vector Institute for Artificial Intelligence, Toronto, Canada}

\icmlcorrespondingauthor{Marc-Etienne Brunet}{mebrunet@cs.toronto.edu}

\icmlkeywords{Machine Learning, ICML}

\vskip 0.3in
]



\printAffiliationsAndNotice{}  

\begin{abstract}
Popular word embedding algorithms exhibit stereotypical biases, such as gender bias.
The widespread use of these algorithms in machine learning systems can thus amplify stereotypes in important contexts.
Although some methods have been developed to mitigate this problem, how word embedding biases arise during training is poorly understood.
In this work, we develop a technique to address this question. Given a word embedding, our method reveals how perturbing the training corpus would affect the resulting embedding bias.
By tracing the origins of word embedding bias back to the original training documents, one can identify subsets of documents whose removal would most reduce bias.
We demonstrate our methodology on Wikipedia and New York Times corpora, and find it to be very accurate.
\end{abstract}

\section{Introduction}
As machine learning algorithms play ever-increasing roles in our lives, there are ever-increasing risks for these algorithms to be systematically biased \cite{Coref, shopping, kleinbergfairness, dworkfairness, hardtequality}. An ongoing research effort is showing that machine learning systems can not only reflect human biases in the data they learn from, but also magnify these biases when deployed in practice~\cite{sweeney2013discrimination}. With algorithms aiding critical decisions ranging from medical diagnoses to hiring decisions, it is important to understand how these biases are learned from data.

In recent work, researchers have uncovered an illuminating example of bias in machine learning systems: Popular word embedding methods such as word2vec~\cite{Word2Vec} and GloVe~\cite{GloVe} acquire stereotypical human biases from the text data they are trained on. For example, they disproportionately associate male terms with science terms, and female terms with art terms~\cite{MachineBias,WEAT}. Deploying these word embedding algorithms in practice, for example in automated translation systems or as hiring aids, thus runs the serious risk of perpetuating problematic biases in important societal contexts. This problem is especially pernicious because these biases can be difficult to detect---for example, word embeddings were in broad industrial use before their stereotypical biases were discovered.

Although the existence of these biases is now established, their \emph{origins}---how biases are learned from training data---are poorly understood. 
Ideally, we would like to be able to ascribe how much of the overall embedding bias is due to any particular small subset of the training corpus---for example, an author or single document. 
Na{\"i}vely, this could be done directly by removing the document in question, re-training an embedding on the perturbed corpus, then comparing the bias of the original embedding with the bias of the retrained embedding. 
The change in bias resulting from this perturbation could then be interpreted as the document's contribution to the overall bias.
But this approach comes at a prohibitive computational cost; completely retraining the embedding for each document is clearly infeasible.

In this work, we develop an efficient and accurate method for solving this problem. 
%
Given a word embedding trained on some corpus, and a metric to evaluate bias, our method approximates how removing a small part of the training corpus would affect the resulting bias.
We decompose this problem into two main subproblems: measuring how perturbing the training data changes the learned word embedding; and measuring how changing the word embedding affects its bias. 
Our central technical contributions solve the former subproblem (the latter is straightforward for many bias measures). 
Our method provides a highly efficient way of understanding the impact of \textit{every} document in a training corpus on the overall bias of a word embedding; therefore, we can rapidly identify the most bias-influencing documents in the training corpus. 
These documents may be used to manipulate the word embedding's bias through highly selective pruning of the training corpus, or they may be analyzed in conjunction with metadata to identify particularly biased subsets of the training data.


We demonstrate the accuracy of our technique with experimental results on both a simplified corpus of Wikipedia articles in broad use~\cite{Simplewiki}, and on a corpus of New York Times articles from 1987--2007~\cite{NewYorkTimes}. 
Across a range of experiments, we find that our method's predictions of how perturbing the input corpus will affect the bias of the embedding are extremely accurate. We study whether our results transfer across embedding methods and bias metrics, and show that our method is much more efficient at identifying bias-inducing documents than other approaches.
We also investigate the qualitative properties of the influential documents surfaced by our method. Our results shed light on how bias is distributed throughout the documents in the training corpora, as well as expose interesting underlying issues in a popular bias metric.

\section{Related Work}
Word embeddings are compact vector representations of words learned from a training corpus, and are actively deployed in a number of domains. They not only preserve statistical relationships present in the training data, generally placing commonly co-occurring words close to each other, but they also preserve higher-order syntactic and semantic structure, capturing relationships such as \textit{Madrid} is to \textit{Spain} as \textit{Paris} is to \textit{France}, and \textit{Man} is to \textit{King} as \textit{Woman} is to \textit{Queen} \cite{Mik13}. However, they have been shown to also preserve problematic relationships in the training data, such as \textit{Man} is to \textit{Computer Programmer} as \textit{Woman} is to \textit{Homemaker} \cite{Homemaker}.

A recent line of work has begun to develop measures to document these biases as well as algorithms to correct for them. 
\citet{WEAT} introduced the Word Embedding Association Test (WEAT) and used it to show that word embeddings trained on large public corpora (e.g., Wikipedia, Google News) consistently replicate the known human biases measured by the Implicit Association Test \cite{IAT98}.
For example, female terms (e.g., ``her'', ``she'', ``woman'') are closer to family and arts terms than they are to career and math terms, whereas the reverse is true for male terms. 
\citet{Homemaker} developed algorithms to de-bias word embeddings so that problematic relationships are no longer preserved, but unproblematic relationships remain.
We build upon this line of work by developing a methodology to understand the sources of these biases in word embeddings.

Stereotypical biases have been found in other machine learning settings as well. Common training datasets for multilabel object classification and visual semantic role labeling contain gender bias and, moreover, models trained on these biased datasets exhibit greater gender bias than the training datasets \cite{shopping}. Other types of bias, such as racial bias, have also been shown to exist in machine learning applications \cite{MachineBias}.

Recently, \citet{Influence} proposed a methodology for using influence functions, a technique from robust statistics, to explain the predictions of a black-box model by tracing the learned state of a model back to individual training examples \cite{InfluenceOriginal}. 
Influence functions allow us to efficiently approximate the effect on model parameters of perturbing a training data point.
Other efforts to increase the explainability of machine learning models have largely focused on providing visual or textual information to the user as justification for classification or reinforcement learning decisions \cite{visualexp2,visualexp,exprobots}.

\section{Background}


\subsection{The GloVe word embedding algorithm} 
Learning a GloVe \cite{GloVe} embedding from a tokenized corpus and a fixed vocabulary of size $V$ is done in two steps. 
First, a sparse co-occurrence matrix $X \in \Real ^{V \times V}$ is extracted from the corpus, where each entry $X_{ij}$ represents a weighted count of the number of times word $j$ occurs in the context of word $i$.
Gradient-based optimization is then used to learn the optimal embedding parameters $w^*$, $u^*$, $b^*$, and $c^*$ which minimize the loss:
\begin{align}\label{eq:glove}
\begin{split}
J(X, w, u, b, c) &= \\
\sum_{i=1}^V \sum_{j=1}^V f(&X_{ij}) (w_i ^T u_j + b_i + c_j - \log X_{ij})^2
\end{split}
\end{align}
where $w_i \in \Real^D$ is the vector representation (embedding) of the $i$th word in the vocabulary, $1 \leq i \leq V$. 
The embedding dimension $D$ is commonly chosen to be between 100 and 500. 
The set of $u_j \in \Real^{D}$ represent the ``context'' word vectors\footnote{When the context window is symmetric, the two sets of vectors are equivalent and differ only based on their initializations.}.
Parameters $b_i$ and $c_j$ represent the bias terms for $w_i$ and $u_j$, respectively.
The weighting function $f(x) = \min((x/x_{max})^\alpha, 1)$ is used to attribute more importance to common word co-occurrences.
The original authors of GloVe used $x_{max} = 100$ and found good performance with $\alpha = 0.75$.
We refer to the final learned emebedding as $w^* =\{w^*_i\}$ throughout.


\subsection{Influence Functions} \label{Influence}
Influence functions offer a way to approximate how a model's learned optimal parameters will change if the training data is perturbed.
We summarize the theory here.

Let $R(z, \theta)$ be a convex scalar loss function for a learning task, with optimal model parameters $\theta^*$ of the form in Equation \eqref{eq:model} below, where $\{z_1, ... , z_n\}$ are the training data points and $L(z_i, \theta)$ is the point-wise loss.
\begin{align}\label{eq:model}
R(z, \theta) = \frac{1}{n} \sum_{i=1}^n L(z_i, \theta) && \theta^* = \argmin_{\theta} R(z, \theta)
\end{align}
We would like to determine how the optimal parameters $\theta^*$ would change if we perturbed a small subset of points in the training set; i.e., when $z_k \rightarrow \tilde{z}_k$ for all $k$ in the set of perturbed indices $\delta$. It can be shown that the perturbed optimal parameters, which we denote $\tilde{\theta}$, can be written as:
\begin{align}\label{eq:influence}
\tilde{\theta} \approx \theta^* - \frac{1}{n} H_{\theta^*} ^{-1} \sum_{k\in\delta}\left[\nabla_{\theta} L(\tilde{z}_k, \theta^*) - \nabla_{\theta} L(z_k, \theta^*)\right]
\end{align}
where $H_{\theta^*} = \frac{1}{n}\sum_{i=1}^n \nabla_{\theta}^2 L(z_i,\theta^*)$ is the Hessian of the total loss, and it is assumed $|\delta| \ll n$. 
Note that we have extended the equations presented by \citet{Influence} to address multiple perturbations.
This is explained in the supplemental materials. 

\subsection{The Word Embedding Association Test}\label{SecWEAT}




The \textit{Word Embedding Association Test} (WEAT) measures bias in word embeddings \cite{WEAT}.
It considers two equal-sized sets $\mathcal{S}$, $\mathcal{T}$ of \textit{target words}, such as $\mathcal{S}=\{$\textit{math, algebra, geometry, calculus}\} and $\mathcal{T}=\{$\textit{poetry, literature, symphony, sculpture}\}, and two sets $\mathcal{A}$, $\mathcal{B}$ of \textit{attribute words}, such as $\mathcal{A}=\{$\textit{male, man, boy, brother, he}\} and $\mathcal{B}=\{$\textit{female, woman, girl, sister, she}\}.

The similarity of words $a$ and $b$ in word embedding $w$ is measured by the cosine similarity of their vectors, $cos(w_a, w_b)$.
The differential association of word $c$ with the word sets $\mathcal{A}$ and $\mathcal{B}$ is measured with:
\begin{align*}
g(c, \mathcal{A}, \mathcal{B}, w) = \substack{\text{mean} \\ a \in \mathcal{A}}\ cos(w_c, w_a) -  \substack{\text{mean} \\ b \in \mathcal{B}}\ cos(w_c, w_b)
\end{align*}
For a given $\{\mathcal{S}, \mathcal{T}, \mathcal{A}, \mathcal{B}\}$, the \textit{effect size} through which we measure bias is:
\begin{align}
B_{\text{weat}}(w) = \frac{ \substack{\text{mean} \\ s \in \mathcal{S}}\ g(s, \mathcal{A}, \mathcal{B}, w) - \substack{\text{mean} \\ t \in \mathcal{T}}\ g(t, \mathcal{A}, \mathcal{B}, w)}{ \substack{\text{std-dev} \\ c \in \mathcal{S} \cup \mathcal{T}}\ g(c, \mathcal{A}, \mathcal{B}, w)}
\end{align}
Where \textit{mean} and \textit{std-dev} refer to the arithmetic mean and the sample standard deviation respectively. 
Note that $B_{\text{weat}}$ only depends on the set of word vectors $\{w_i |\ i \in \mathcal{S} \cup \mathcal{T} \cup \mathcal{A} \cup \mathcal{B}\}.$

\section{Methodology}

Our technical contributions are twofold. First, we formalize the problem of understanding bias in word embeddings, introducing the concepts of \textit{differential bias} and \textit{bias gradient}. Then, we show how the differential bias can be approximated in word embeddings trained using the GloVe algorithm. We address how to approximate the bias gradient in GloVe in the supplemental material.

\subsection{Formalizing the Problem}
\xhdr{Differential Bias}
Let $w=\{w_1, w_2, ... ,w_V\}$, $w_i \in \Real^D$ be a word embedding learned on a corpus $C$.
Let $B(w)$ denote any bias metric that takes as input a word embedding and outputs a scalar.
Consider a partition of the corpus into many small parts (e.g. paragraphs, documents), and let $p$ be one of those parts.
Let $\tilde{w}$ be the word embedding learned from the perturbed corpus $\tilde{C} = C \setminus p$. 
We define the \textit{differential bias} of part $p \subset C$ to be:
\begin{align}\label{eq:diff_bias}
\Delta_{p} B = B(w) - B(\tilde{w})
\end{align}
Which is the incremental contribution of part $p$ to the total bias. This value decomposes the total bias, enabling a wide range of analyses (e.g., studying bias across metadata associated with each part).

It is natural to think of $C$ as a collection of individual documents, and think of $p$ as a single document. 
Since a word embedding is generally trained on a corpus consisting of a large set of individual documents (e.g., websites, newspaper articles, Wikipedia entries), we use this framing throughout our analysis. 
Nonetheless, we note that the unit of analysis can take an arbitrary size (e.g., paragraphs, sets of documents), provided that only a relatively small portion of the corpus is removed. 
Thus our methodology allows an analyst to study how bias varies across documents, groups of documents, or whichever grouping is best suited to the domain.

\xhdr{Co-occurrence perturbations}
Several word embedding algorithms, including GloVe, operate on a co-occurrence matrix rather than directly on the corpus.
The co-occurrence matrix $X$ is a function of the corpus $C$, and can be viewed as being constructed additively from the co-occurrence matrices of the $n$ individual documents in the corpus, where $X^{(k)}$ is the co-occurrence matrix for document $k$.
In this manner, we can view $X$ as $X = \sum_{k=1}^n X^{(k)}$.
We then define $\tilde{X}$ as the co-occurrence matrix constructed from the perturbed corpus $\tilde{C}$.
If $\tilde{C}$ is obtained by omitting document $k$, we have $\tilde{X} = X - X^{(k)}$.

\xhdr{Bias Gradient}
If a word embedding $w$ is (or can be approximated by) a differentiable function of the co-occurrence matrix $X$, and the bias metric $B(w)$ is also differentiable, we can consider the \textit{bias gradient}:
\begin{align}
\label{eq:bias_grad}
\begin{split}
\nabla_{X} B(w(X)) &=  \nabla_{w} B(w) \nabla_{X} w(X)
\end{split}
\end{align}
Where the above equality is obtained using the chain rule.

The bias gradient has the same dimension as the co-occurrence matrix $X$.
While $V \times V$ is a daunting size, if the bias metric is only affected by a small subset of the words in the vocabulary, as is the case with the WEAT bias metric, the gradient will be very sparse. 
It may then be feasible to compute and study.
Since it ``points'' in the direction of maximal bias increase, it provides insight into the co-occurrences most affecting bias.

The bias gradient can also be used to linearly approximate how the bias will change due to a small perturbation of $X$. It can therefore be used to approximate the differential bias of document $k$. Again letting $\tilde{X} = X - X^{(k)}$, we start from a first order Taylor approximation
of $B(w(\tilde{X}))$ around $X$:
\begin{align*}
B(w(\tilde{X})) \approx B(w(X)) - \nabla_X B(w(X)) \cdot X^{(k)}
\end{align*}
We then rearrange, and apply the chain rule, obtaining:
\begin{align*}
B(w(X)) - B(w(\tilde{X})) \approx \nabla_{w} B(w) \nabla_{X} w(X) \cdot X^{(k)}
\end{align*}
Where $w(\tilde{X})$ is equivalent to the $\tilde{w}$ of Equation \eqref{eq:diff_bias}.


\subsection{Computing the Differential Bias for GloVe}\label{sec:diff_bias}
The naive way to compute the differential bias for a document is to simply remove the document from the corpus and retrain the embedding.
However, if we wish to learn the differential bias, of every document in the corpus, this approach is clearly computationally infeasible.
Instead of computing the perturbed embedding $\tilde{w}$ directly, we calculate an approximation of it by applying a tailored version of influence functions. Generally, influence functions require the use of $H_{\theta^*} ^{-1}$, as in Equation \eqref{eq:influence}.
In the case of GloVe this would be a $2V(D+1)$ by $2V(D+1)$ matrix, which would be much too large to work with. 

\xhdr{The need for a new method}
To overcome the computational barrier of using influence functions in large models, \citet{Influence} use the LiSSA algorithm \cite{LiSSA} to efficiently compute inverse Hessian vector products. 
They compute influence in roughly $\BigO{np}$ time, where $p$ is the number of model parameters and $n$ is the number of training examples. 
However, our analysis and initial experimentation showed that this method would still be too slow for our needs. 
In a typical setup, GloVe simply has too many model parameters ($2V(D+1)$), and most corpora of interest cause $n$ to be too large. 
One of our principal contributions is a simplifying assumption about the behavior of the GloVe loss function around the learned embedding $w^*$. This simplification causes the Hessian of the loss to be block diagonal, allowing for the rapid and accurate approximation of the differential bias for every document in a corpus.

\xhdr{Tractably approximating influence functions}
To approximate $\tilde{w}$ using influence functions, we must apply Equation \eqref{eq:influence} to the GloVe loss function from Equation \eqref{eq:glove}.
In doing so, we make a simplifying assumption, treating the  GloVe parameters $u$, $b$, and $c$ as constants throughout the analysis.
As a result, the parameters $\theta$ consist only of $w$ (i.e., $u$, $b$, and $c$ are excluded from $\theta$).
The number of points $n$ is $V$, and the training points $z = \{ z_i\}$ are in our case $X = \{ X_{i}\}$, where $X_i$ refers to the $i$th row of the co-occurrence matrix (not to be confused with the co-occurrence matrix of the $i$th document, denoted as $X^{(i)}$). With these variables mapped over, the point-wise loss function for GloVe becomes:
\begin{align*}
L(X_i, w) =  \sum_{j=1}^V V f(X_{ij}) (w_i ^T u_j + b_i + c_j - \log X_{ij})^2
\end{align*}
and the total loss is then $J(X, w) = \frac{1}{V}\sum_{i=1}^{V}L(X_i, w)$, now in the form of Equation \eqref{eq:model}.

Note that our embedding $w^*$ is still learned through dynamic updates of all of the parameters. It is only in deriving this influence function-based approximation for $\tilde{w}$ that we treat $u$, $b$, and $c$ as constants.

In order to use Equation \eqref{eq:influence} to approximate $\tilde{w}$ we need an expression for the gradient with respect to $w$ of the point-wise loss, $\nabla_w L(X_i, w)$, as well as the Hessian of the total loss, $H_w$. We derive these here, starting with the gradient.

Recall that $w=\{w_1, w_2, ... ,w_V\}$, $w_k \in \Real^D$.
We observe that $L(X_i, w)$ depends only on $w_i$, $u$, $b_i$, and $c$; no word vector $w_k$ with $k \neq i$ is needed to compute the point-wise loss at $X_i$. 
Because of this, $\nabla_w L(X_i, w)$, the gradient with respect to $w$ (a vector in $\Real^{VD}$), will have only $D$ non-zero entries.
These non-zero entries are the entries in $\nabla_{w_i} L(X_i, w)$, the gradient of the point-wise loss function at $X_i$ with respect to only word vector $w_i$. Visually, this is as follows:
\begin{align}
\label{eg:glove_grad}
\begin{split}
\nabla_w L&(X_i, w) = \bigg(\underbrace{\overbrace{0, ..., 0}^{D(i-1)} \,, \overbrace{\nabla_{w_i}L(X_i, w)}^{D}, \, \overbrace{0, ..., 0}^{D(V-i)}}_{VD\, \text{dimensions}} \bigg)
\end{split}
\end{align}
where the D-dimensional vector given by $\nabla_{w_i}L(X_i, w)$ is:
\begin{align*}
\sum_{j=1}^V 2V f(X_{ij}) (w_i ^T u_j + b_i + c_j - \log X_{ij})\,u_j
\end{align*}

From Equation \eqref{eg:glove_grad}, we see that the Hessian of the point-wise loss with respect to $w$, $\nabla_w^2 L(X_i, w)$ (a $VD \times VD$-dimensional matrix), is extremely sparse, consisting of only a single $D \times D$ block in the $i$th diagonal block position. As a result, the Hessian of the total loss, $H_w = \frac{1}{V}\sum_{i=1}^{V}\nabla_{w}^2 L(X_i, w)$ (also a $VD \times VD$ matrix), is block diagonal, with $V$ blocks of dimension $D \times D$. Each $D \times D$ diagonal block is given by:
\begin{align*}
H_{w_i} = \frac{1}{V} \nabla_{w_i}^2 L(X_i, w) = \sum_{j=1}^{V}2f(X_{ij})u_ju_j^T
\end{align*}
which is the Hessian with respect to only word vector $w_i$ of the point-wise loss at $X_i$.

This block-diagonal structure allows us to solve for each $\tilde{w}_i$ independently. Moreover, $\tilde{w}_i$ will only differ from $w^*_i$ for the tiny fraction of words whose co-occurrences are affected by the removal of the selected document for the corpus perturbation. We can approximate how any word vector will change due to a given corpus perturbation with:
\begin{align}\label{eq:glove_if}
\begin{split}
\tilde{w_i} \approx w^*_i - \frac{1}{V} H_{w_i}^{-1}\big[\nabla_{w_i} L(\tilde{X}_i, w) - \nabla_{w_i}L(X_i, w)\big]
\end{split}
\end{align}
\xhdr{An efficient algorithm} Combining Equation \eqref{eq:glove_if} with Equation \eqref{eq:diff_bias}, we can approximate the differential bias of every document in the corpus.
Notice in Equation \eqref{eq:glove_if} that $\tilde{w}_i = w_i^*$ for all $i$ where $\tilde{X}_i = X_i$.
Also recall that $B_{\text{weat}}$ only depends on a small set of WEAT words $\{\mathcal{S}, \mathcal{T}, \mathcal{A}, \mathcal{B}\}$.
Therefore, when approximating the differential bias for a document, we only need to compute $\tilde{w_i}$ for the WEAT words in that document. 
This is outlined in Algorithm \ref{alg:methodology}.

\begin{algorithm}
   \caption{Approximating Differential Bias}
   \label{alg:methodology}
\begin{algorithmic}
    \INPUT \textit{Co-occ Matrix}: $X$, \textit{WEAT words}: $\{\mathcal{S}, \mathcal{T}, \mathcal{A}, \mathcal{B}\}$
    \STATE {$w^*$, $u^*$, $b^*$, $c^*$ = GloVe($X$)} \textit{\# Train embedding}
    \FOR { $\text{doc}$ {\bfseries in} corpus }
    \STATE $\tilde{X} = X - X^{(k)}$ \textit{\# Subtract coocs from doc k}
    \FOR {word $i$ {\bfseries in} $\text{doc} \cap (\mathcal{S} \cup \mathcal{T} \cup \mathcal{A} \cup \mathcal{B})$}
    \STATE \textit{\# Only need change in WEAT word vectors}
    \STATE $\tilde{w}_i = w^*_i - \frac{1}{V} H_{w_i}^{-1} \big[\nabla_{w_i} L(\tilde{X}_i, w) - \nabla_{w_i}L(X_i, w)\big]$
    \ENDFOR
    \STATE $\Delta_{\text{doc}} B \approx B_{\text{weat}}(w^*) - B_{\text{weat}}(\tilde{w})$
    \ENDFOR
\end{algorithmic}
\end{algorithm}

\section{Experimentation}
Our experimentation has several objectives. 
First, we test the accuracy of our differential bias approximation.
We then compare our method to a simpler count-based baseline. 
We also test whether the documents which we identify as bias influencing in GloVe embeddings affect bias in word2vec.
Finally, we investigate the qualitative properties of the influential documents surfaced by our method. 
Our results shed light on how bias is distributed throughout the documents in the training corpora, and expose interesting underlying issues in the WEAT bias metric.



\subsection{Experimental Setup}

\xhdr{Choice of corpus and hyperparameters}
We use two corpora in our experiments, each with a different set of GloVe hyperparameters.
This first setup consists of a corpus constructed from a Simple English Wikipedia dump (2017-11-03)~\cite{Simplewiki} using 75-dimensional word vectors.
These dimensions are small by the standards of a typical word embedding, but sufficient to start capturing syntactic and semantic meaning. Performance on the TOP-1 analogies test shipped with the GloVe code base was around 35\%, lower than state-of-the-art performance but still clearly capturing significant meaning.

Our second setup is more representative of the academic and commercial contexts in which our technique could be applied. The corpus is constructed from 20 years of New York Times (NYT) articles \cite{NewYorkTimes}, using 200-dimensional vectors. The TOP-1 analogy performance is approximately 54\%. The details of these two configurations are tabulated in the supplemental material.

\xhdr{Choice of experimental bias metric}
Throughout our experiments, we consider the \emph{effect size} of two different WEAT biases as presented by \citet{WEAT}.
Recall that these metrics have been shown to correlate with known human biases as measured by the Implicit Association Test.
In WEAT1, the target word sets are \emph{science} and \emph{arts} terms, while the attribute word sets are \emph{male} and \emph{female} terms.
In WEAT2, the target word sets are \emph{musical instruments} and \emph{weapons}, while the attribute word sets are \emph{pleasant} and \emph{unpleasant} terms.
A full list of the words in these sets can be found in the supplemental material. They are summarized in Table \ref{tab:WEAT}.
These sets were chosen so as to include one societal bias that would be widely viewed as problematic, and another which would be widely viewed as benign. 


                                 

\subsection{Testing the Accuracy of our Method}
\xhdr{Experimental Methodology}
To test the accuracy of our methodology, ideally we would simply remove a single document from a word embedding's corpus, train a new embedding, and compare the change in bias with our differential bias approximation.
However, the cosine similarities between small sets of word vectors in two word embeddings trained on the same corpus can differ considerably simply because of the stochastic nature of the optimization \cite{stability}.
As a result, the WEAT biases vary between training runs. 
The effect of removing a single document, which is near zero for a typical document, is hidden in this variation.
Fixing the random seed is not a practical approach.
Many popular word embedding implementations also require limiting training to a single thread to fully eliminate randomness.
This would make experimentation prohibitively slow.

In order to obtain measurable changes, we instead remove sets of documents, resulting in larger corpus perturbations.
Accuracy is assessed by comparing our method's predictions to the actual change in bias measured when each document set is removed from the corpus and a new embedding is trained on this perturbed corpus. 
Furthermore, we make all predictions and assessments using several embeddings, each trained with the same hyperparameters, but differing in their random seeds.

\begin{figure}[t]
    \centering
    \includegraphics[width=65mm]{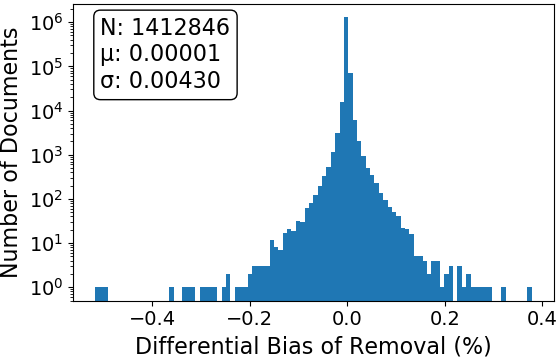}
    \caption{Histogram of the approximated differential bias of removal for every document in our NYT setup, considering WEAT1, measured in percent change from the baseline mean.}
    \label{fig:histograms}
\end{figure}

We construct three types of perturbation sets: \emph{increase}, \emph{random}, and \emph{decrease}.
The targeted (increase, decrease) perturbation sets are constructed from the documents whose removals were predicted (by our method) to cause the greatest differential bias, e.g., the documents located in the tails of the histograms in Figure \ref{fig:histograms}.
The random perturbation sets are simply documents chosen from the corpus uniformly at random.
For a more detailed description, please refer to the supplemental material. 
Most of the code used in the experimentation has been made available online\footnote{Code at https://github.com/mebrunet/understanding-bias}. 


\begin{table}
\caption{WEAT Target and Attribute Sets}
\label{tab:WEAT}
\centering
\begin{tabular}{ l l  c  c }
\hline
    & & WEAT1 & WEAT2\\ \hline
    \multirow{2}{*}{Target Sets}    & $\mathcal{S}$ & science & instruments \\
                                    & $\mathcal{T}$ & arts & weapons \\ \hline
    \multirow{2}{*}{Attribute Sets} & $\mathcal{A}$ & male & pleasant \\
                                    & $\mathcal{B}$ & female & unpleasant \\
                                 
\end{tabular}
\end{table}


\xhdr{Experimental Results}\label{results}
Here we present a subset of our experimental results, principally from NYT WEAT1 (science vs. arts). 
Complete sets of results from the four configurations (\{NYT, Wiki\} $\times$ \{WEAT1, WEAT2\}) can be found in the supplemental materials. 

The baseline WEAT effect sizes ($\pm$ 1 std. dev.) are shown in Table \ref{tab:baseline}. It is worth noting that the WEAT2 (weapons vs. instruments) bias was not significant in our Wiki setup.
However, our analysis does not require that the bias under consideration fall within any particular range of values.

\begin{table}
    \caption{Baseline WEAT Effect Sizes}
    \smallskip
    \label{tab:baseline}
    \centering
    \begin{tabular}{ l  c  c }
    \hline
         & WEAT1                 & WEAT2 \\ \hline
    Wiki & 0.957 ($\pm$ 0.150) & 0.108 ($\pm$ 0.213)\\
    NYT  & 1.14, ($\pm$ 0.124) & 1.32, ($\pm$ 0.056)\\
    \end{tabular}
\end{table}

A histogram of the differential bias of removal for each document in our NYT setup (WEAT1) can be seen in Figure~\ref{fig:histograms}. Notice the log scale on the vertical axis, and how the vast majority of documents are predicted to have a very small impact on the differential bias.

\begin{figure}
  \centering
  \includegraphics[width=50mm]{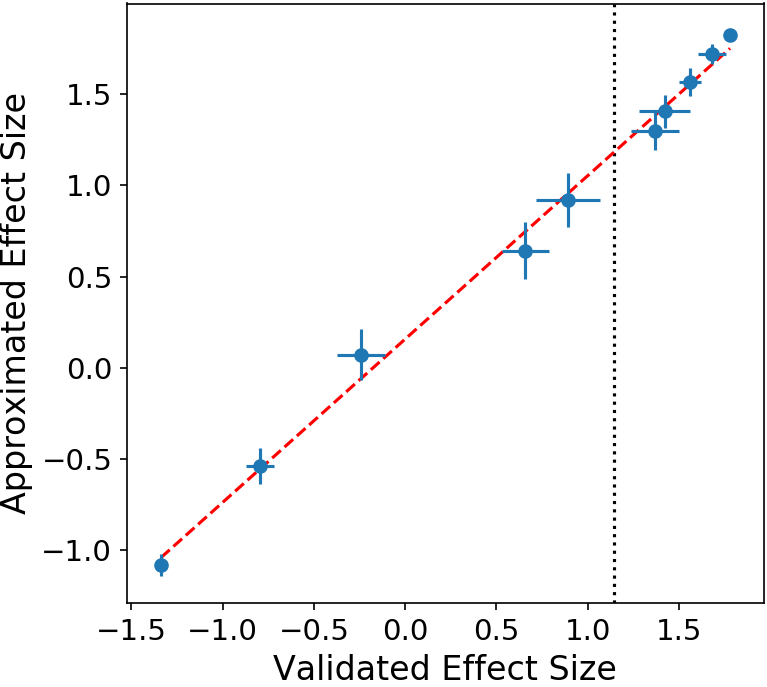}
  \caption{Approximated and ground truth WEAT bias effect size due to the removal of various perturbation sets for our NYT corpus, considering WEAT1. 
  Each point describes the mean effect size of one set; error bars depict one standard deviation; 
  the baseline (unperturbed) mean is shown with a vertical dotted line.}
  \label{fig:means}
\end{figure}

We assess the accuracy of our approximations by measuring how they correlate with the ground truth change in bias (as measured by retraining the embedding after removing a subset of the training corpus). Recall these ground truth changes are obtained using several retraining runs with different random seeds. We find extremely strong correlations ($r^2 \geq 0.985$) in every configuration, for example Figure~\ref{fig:means}.

\begin{figure}
  \centering
  \includegraphics[width=65mm]{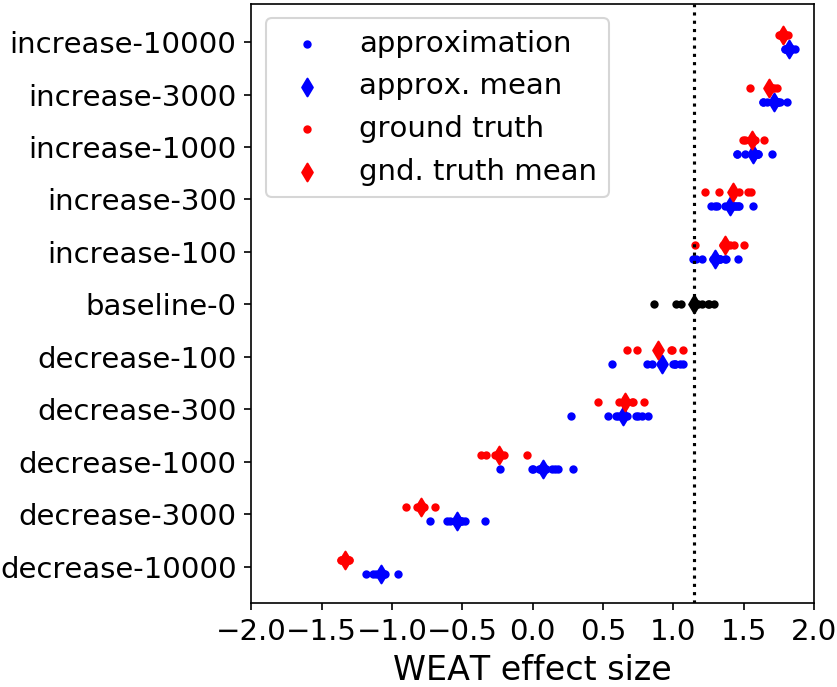}
  \caption{Approximated and ground truth differential bias of removal for every perturbation set. Results for different perturbation sets arranged vertically, named as \textit{type - size (number of documents removed)}. (NYT - WEAT1)}
  \label{fig:targeted}
\end{figure}

We further compare our approximations to the ground truth in Figure \ref{fig:targeted}.
We see that while our approximations underestimate the magnitude of the change in effect size when the perturbation causes the bias to invert, relative ranking is nonetheless preserved.
There was no apparent change in the TOP-1 analogy performance of the perturbed embeddings.

We ran a Welch's t-test comparing the perturbed embeddings' biases with the baseline biases measured in the original (unperturbed) embeddings.
For 36 random perturbation sets, only 2 differed significantly ($p < 0.05$) from the baseline.
Both of these sets were perturbations of the smaller Wiki corpus and they only caused a significant difference for WEAT2.
This is in strong contrast to the 40 targeted perturbation sets, where only 2 did \emph{not} significantly differ from their respective baselines.
In this case, both were from the smallest (10 document) perturbation sets.



\subsection{Comparison to a PPMI Baseline}
We have shown that our method can be used to identify bias-influencing documents and accurately approximate the impact of their removal, but how does it compare to a more naive, straightforward approach? 
The positive point-wise mutual information (PPMI) matrix is a count-based distributed representation commonly used in natural language processing \cite{PPMI}. 
We compare the WEAT effect size in our NYT GloVe embeddings versus when measured in the corpus' PPMI representation (on 2000 randomly generated word sets).
As expected, there is a clear correlation ($r^2=0.725$). 
It is therefore sensible to use the change in PPMI WEAT effect size to predict how the GloVe WEAT effect size will change.

A change in the PPMI representation due to a co-occurrence perturbation (e.g.\ document removal) can be computed rapidly.
This allows us to scan the whole corpus for the most bias influencing documents.
However, we find that the documents identified in this way have a much smaller impact on the bias than those identified by our method. 
For example in our Wiki setup (WEAT1) removing the 10 documents identified as most bias increasing by the PPMI method reduced the WEAT effect size by 4\%.
In contrast, the 10 identified by our method reduced it by 40\%.
Further comparisons are tabulated in the supplemental material. 


\subsection{Impact on Word2Vec and Other Bias Metrics}
The documents identified as influential by our method clearly have a strong impact on the WEAT effect size in GloVe embeddings. Here we explore how those same documents impact the bias in word2vec embeddings, as well as other bias metrics. 

\begin{figure}
  \centering
  \includegraphics[width=65mm]{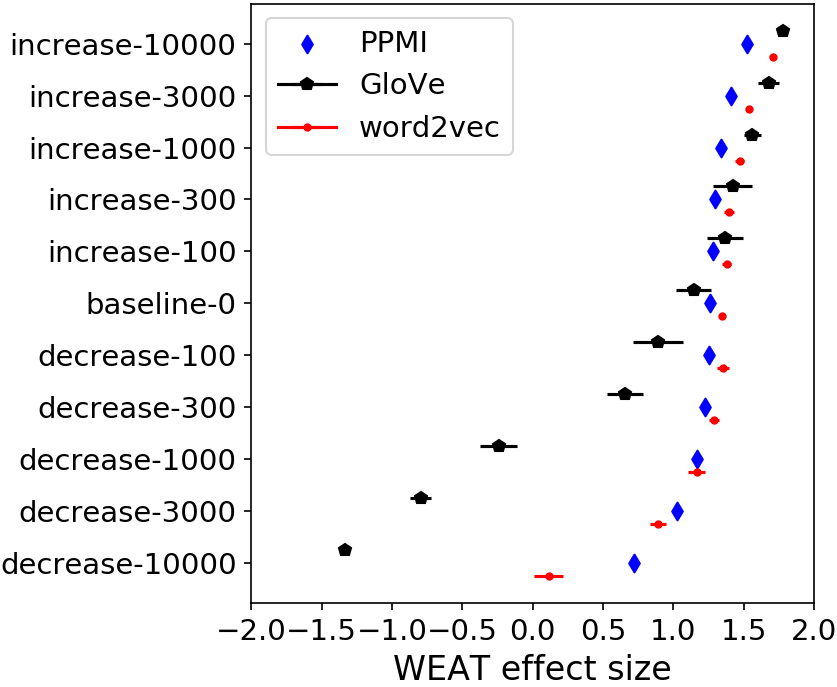}
  \caption{The effects of removing the different perturbation sets (most impactful documents as identified by our method) on the WEAT bias in: our GloVe embeddings, the PPMI representation, and word2vec embeddings with comparable hyper-parameters; error bars represent one standard deviation. (NYT - WEAT1)}
  \label{fig:transfer}
\end{figure}

We start by training five word2vec emebeddings with comparable hyperparameters\footnote{We use a CBOW architecture with the same vocabulary, vector dimensions, and window size as our GloVe embeddings.} for each perturbation set, and measure how their removals affect the bias. 
Figure \ref{fig:transfer} shows how the WEAT effect size changes in GloVe, the PPMI, and word2vec for each set (NYT-WEAT1). 
We see that while the response is weaker, both the PPMI representation and the word2vec embeddings show a clear change in effect size due to the perturbations.
For example, the baseline WEAT effect size in word2vec is $1.35$ in the unperturbed corpus, but after removing \textit{decrease-10000} (the 10k most bias contributing documents for GloVe), the effect size drops to $0.11$.
This means we have nearly neutralized the bias in word2vec through the removal of less than $1\%$ of the corpus (and there is no significant change in TOP-1 analogy performance).

\begin{figure}
  \centering
  \includegraphics[width=65mm]{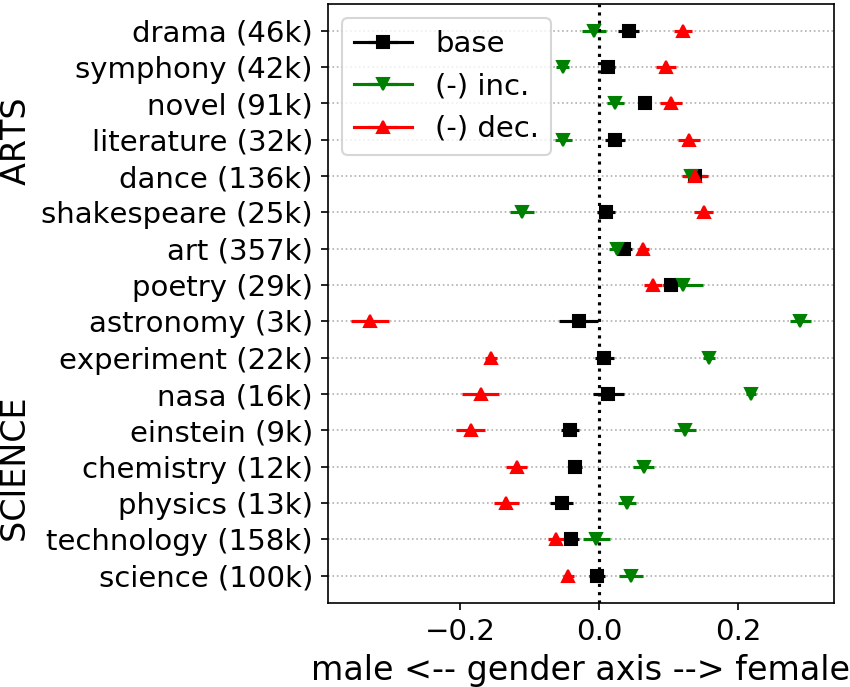}
  \caption{The effect of \textit{removing} the 10k most bias increasing and bias decreasing documents as identified by our method on the projection of the target words onto the gender axis vs. unperturbed corpus (base); error bars show one std dev; corpus word frequency noted in parentheses. (NYT - WEAT1)
}
  \label{fig:direct_bias}
\end{figure}

We also see a change as measured by other bias metrics in our perturbed GloVe embeddings. 
The metric proposed by \citet{Homemaker} involves computing a single dimensional gender subspace using a definitional sets of words. 
One can then project test words onto this axis and measure how the embedding implicitly genders them. 
We explore this in our NYT setup by using the WEAT 1 attribute word sets (male, female) to construct a gender axis, then projecting the target words (science, arts) onto it. 
In Figure \ref{fig:direct_bias} we show the baseline projections and compare them to the projections after having removed the 10k most bias increasing and bias decreasing documents.
We see a strong response to the perturbations in the expected directions.

\subsection{Qualitative Analysis}
We've demonstrated that removing the most influential documents identified by our methodology significantly impacts the WEAT, a metric that has been shown to correlate with known human biases.
But can the \textit{semantic content} of these documents be intuitively understood to affect bias?



We comment here on the 50 most bias influencing documents in the New York Times corpus, considering the WEAT 1 bias metric (\{\textit{male}, \textit{female}\}, \{\textit{science}, \textit{arts}\}).
This list is included in the supplemental materials. 
We indeed found that most of these documents could be readily understood to affect the bias in the expected semantic sense. 
For example, the second most bias \textit{decreasing} document is entitled \textit{``For Women in Astronomy, a Glass Ceiling in the Sky''}, which investigates the pay and recognition gap in astronomy. 
Many of the other bias decreasing documents included interviews with female doctors or scientists.

Correspondingly, the most bias \textit{increasing} documents consisted mainly of articles describing the work of male engineers and scientists.
There were several obituary entries detailing the scientific accomplishments of men, e.g., \textit{``Kaj Aage Strand, 93, Astronomer At the U.S. Naval Observatory''}.
Perhaps the most self-evident example was an article entitled \textit{``60 New Members Elected to Academy of Sciences''}, a list of almost exclusively male scientists receiving awards. 

There were, however, a few examples of articles that seemed like their semantic content should affect the bias inversely to how they were categorized.
For example, an article entitled  \textit{``The Guide''}, a guide to events in Long Island, mentions that the group Woman in Science would be hosting an astronomy event, but nonetheless \textit{increases} the bias. 
Only 2 or 3 documents seemed altogether unrelated to the bias' theme.

Surprisingly, some of the most bias influencing articles contained none of the science or arts WEAT terms explicitly, only synonyms (and some of the male or female terms).
This shows that the impact of secondary co-occurrences can be very strong.
A naive approach to understanding bias may only consider co-occurrences between WEAT words, but our method shows that this would miss some of the most bias influencing documents in the corpus.

Importantly, we also noticed a large portion of the most bias influencing documents dealt with \textit{astronomy} or contained \textit{hers}, the rarest words their respective WEAT subsets.
Upon further investigation, we found that the log of a word's frequency is correlated with the extent to which its relative position (among WEAT words) is affected by the perturbation sets $(r^2 = 0.828)$.
This can be seen in Figure \ref{fig:direct_bias}.
Not surprisingly, our results indicate that the embedded representations of rare words are more sensitive to corpus perturbations.
However, this leaves the WEAT metric vulnerable to exploitation through the manipulation of rarer words.
The WEAT effect size is an average of cosine-similarities between the embedded representations of four subsets of words. 
A handful of well chosen documents can significantly alter the embeddings of a few rare words in those subsets.
Therefore documents containing the rare words can have a disproportionate impact on the metric. 
This weakness helps explain how removing a mere $0.07\%$ of articles can reverse the WEAT effect size in the New York Times, as is shown in Figure \ref{fig:targeted}, \textit{decrease-1000}. 



\section{Conclusion} \label{discussion}
In this work, we introduce the problem of tracing the origins of bias in word embeddings, and we develop and experimentally validate a methodology to solve it. We conceptualize the problem as measuring the resulting change in bias when we remove a training document (or small subset of the training corpus), and interpret this as the amount of bias contributed by the document to the overall embedding bias. 
Computing this naively for each training document would be infeasible.
We develop an efficient approximation of this differential bias using influence functions and apply it to the GloVe word embedding algorithm.
We experimentally validate our approach and find that it very accurately approximates the true change in bias that results from manually removing training documents and retraining.
It performs well on tests using Simple Wikipedia and New York Times corpora and two WEAT bias metrics. 

Our work represents a new approach to understanding how machine learning algorithms learn biases from training data. 
Our methodology could be applied to assess how the bias of a set of texts has evolved over time. For example, using publicly available datasets of newspaper articles or books, one could measure how cultural biases as measured by WEAT or other metrics have evolved over time. 
More broadly, our efficient method
for tracing how perturbations in training data affect changes in the bias of the output is a general idea, and could be applied in many other contexts.
\newpage
\bibliography{main}
\bibliographystyle{icml2019}

\newpage
\onecolumn
\section*{Supplemental Material}
\appendix
\section{Computing the Bias Gradient for GloVe}

The bias gradient $\nabla_{X} B(w(X))$ can be thought of as a $V \times V$ matrix indicating the direction of perturbation of the corpus (co-occurrences) that will result in the maximal change in bias. 
\begin{align*}
\begin{split}
\nabla_{X} B(w(X)) &=  \nabla_{w} B(w) \nabla_{X} w(X) \\
                   &= \sum_{i=1}^{V} \nabla_{w_i} B(w) \nabla_{X} w_i(X)
\end{split}
\end{align*}
Where the first line is obtained through the chain rule, and the second line is a partial expansion of the resulting Jacobian product. Recall $w = \{w_1, w_2, ... w_V\}$, $w_i \in \Real^D$ and $X \in \Real^{V \times V}$.

When the bias metric is only a function of a small subset of the words in the vocabulary, as in the case of WEAT, this can be further simplified to:
\begin{align}
\begin{split}
\label{eq:bias_grad_ext}
\nabla_{X} B(w(X)) &=  \sum_{i \in \mathcal{U}} \nabla_{w_i} B(w) \nabla_{X} w_i(X)
\end{split}
\end{align}
Where $\mathcal{U}$ are the indices of the words used by the bias metric;  $\mathcal{U} = \mathcal{S} \cup \mathcal{T} \cup \mathcal{A} \cup \mathcal{B}$ for WEAT.
For the bias metrics we have explored, the first part of this expression, $\nabla_{w_i} B(w)$, can be efficiently computed through automatic differentiation. The difficulty lies in finding an expression for $\nabla_{X} w_i(X)$. However, in Section 4.2 of the main text we developed an approximation for the learned embedding under corpus (co-occurrence) perturbations in GloVe using influence functions. We can use this same approximation to create an expression for $w_i(X)$ that is differentiable in $X$.

Recall, given the learned optimal GloVe parameters $w^*$, $u^*$, $b^*$, $c^*$, on co-occurrence matrix $X$, we can approximate the word vectors given a small corpus perturbation as:
\begin{align}
\begin{split}
\label{eq:glove_if_2}
\tilde{w_i} \approx w_i^* - \frac{1}{V} H_{w_i}^{-1}\big[\nabla_{w_i} L(\tilde{X}_i, w^*) - \nabla_{w_i}L(X_i, w^*)\big]
\end{split}
\end{align}
Until now, we have been interested in perturbations stemming from the removal of some part of corpus, e.g. document $k$, giving us $\tilde{X} = X - X^{(k)}$. However, the above approximation holds for an (almost) arbitrary co-occurrence perturbation, which we shall denote $Y$. With this change of variable, $\tilde{X} = X - Y$, we can introduce the approximation from Equation \eqref{eq:glove_if_2}:
\begin{align}
\begin{split}
\label{eq:glove_if_grad}
\nabla_{X} w_i(X) &= -\nabla_{Y}w_i(\tilde{X}(Y))\rvert_{Y=0} \\
                &\approx -\nabla_{Y} \Big[w_i^* - \frac{1}{V} H_{w_i}^{-1}\big[\nabla_{w_i} L(\tilde{X}_i(Y), w^*) - \nabla_{w_i}L(X_i, w^*)\big]\Big]\rvert_{Y=0}\\
                &\approx \frac{1}{V} H_{w_i}^{-1} \nabla_{Y} \nabla_{w_i} L(\tilde{X}_i(Y), w^*)\rvert_{Y=0}
\end{split}
\end{align}
Where we have made the dependence on $Y$ in Equation \eqref{eq:glove_if_2} explicit. 
The higher-order jacobian, $\nabla_{Y} \nabla_{w_i} L(\tilde{X}_i(Y), w^*)\rvert_{Y=0}$, can be thought of as a $D \times V \times V$ tensor.
We again note a significant sparsity, since $\tilde{X}_i(Y)$ is only a function of $Y_i$.
Therefore, this tensor is $0$ in all but the $i$th position along one of the $V$ axes. The $D \times V$ ``matrix'' in that non-zero position can be found by computing:
\begin{align*}
\nabla_{Y_i}\sum_{j=1}^V 2V f(X_{ij}-Y_{ij}) \big(w_i ^T u_j + b_i + c_j - \log (X_{ij}-Y_{ij})\big)\,u_j
\end{align*}
evaluated at $Y_{ij} = 0$. Alternatively the Jacobian can simply by obtained using automatic differentiation.

Substituting this result into Equation \eqref{eq:bias_grad_ext}, we get:
\begin{align*}
\begin{split}
\nabla_{X} B(w(X)) &= \sum_{i \in \mathcal{U}} \nabla_{w_i} B(w) \nabla_{X} w_i(X) \\
    &\approx \frac{1}{V}\sum_{i \in \mathcal{U}} \nabla_{w_i} B(w) H_{w_i}^{-1} \nabla_{Y} \nabla_{w_i} L(\tilde{X}_i(Y), w^*)\rvert_{Y=0}
\end{split}
\end{align*}
Which gives us the full approximation of the Bias Gradient in GloVe. 

Note that since $\nabla_{w_i} L(\tilde{X}_i(Y), w^*)$ is not differentiable in $Y$ at $Y=0$ where $X_{ij}=0$, the bias gradient is only defined at non-zero co-occurrences.
This prevents us from using the bias gradient to study corpus additions which create previously unseen word co-occurrences.
However, this does not affect our ability to study arbitrary removals from the corpus, since removals cannot affect a zero-valued co-occurrence.
Of course, nothing limits us from using the bias gradient to also consider additions to the corpus that not change the set of zero co-occurrences.

\section{Experimental Setup}
Table \ref{tab:setup} presents a summary of the corpora and embedding hyperparameters used throughout our experimentation.
We list the complete set of words used in each of the two WEATs below.

\begin{table}[hb]
\caption{Experimental Setups}
\label{tab:setup}
\begin{center}
\begin{tabular}{ l  c  c }
\toprule
                 & \textbf{Wiki}        &  \textbf{NYT} \\ \midrule
\textbf{Corpus} \\
Min. doc. length & 200         & 100 \\
Max. doc. length & 10,000      & 30,000 \\
Num. documents   & 29,344      & 1,412,846 \\
Num. tokens      & 17,033,637  & 975,624,317 \\
\midrule
\textbf{Vocabulary} \\
Token min. count & 15          & 15 \\
Vocabulary size  & 44,806      & 213,687 \\
\midrule
\textbf{GloVe} \\
Context window   & symmetric   & symmetric \\
Window size      & 8           & 8 \\
$\alpha$         & 0.75        & 0.75 \\
$x_{max}$        & 100         & 100 \\
Vector Dimension & 75          & 200 \\
Training epochs  & 300         & 150 \\
\midrule
\textbf{Performance} \\
TOP-1 Analogy    & $35$\%     & $54$\%
\end{tabular}
\end{center}
\end{table}

\begin{center}
\begin{tabular}{ c  c  p{8cm} }
\multicolumn{3}{c}{WEAT 1} \\
\toprule
$\mathcal{S}$ & \textbf{science} & science, technology, physics, chemistry, einstein, nasa, experiment, astronomy \\
\midrule
$\mathcal{T}$ & \textbf{arts} & poetry, art, shakespeare, dance, literature, novel, symphony, drama \\
\midrule
$\mathcal{A}$ & \textbf{male} & male, man, boy, brother, he, him, his, son \\
\midrule
$\mathcal{B}$ & \textbf{female} & female, woman, girl, sister, she, her, hers, daughter \\
\midrule
\end{tabular}
\end{center}
\bigskip

\bigskip
\begin{center}
\begin{tabular}{ c c p{8cm} }
\multicolumn{3}{c}{WEAT 2} \\
\toprule
$\mathcal{S}$ & \textbf{instruments} & bagpipe, cello, guitar, lute, trombone, banjo, clarinet, harmonica, mandolin, trumpet, bassoon, drum, harp, oboe, tuba, bell, fiddle, harpsichord, piano, viola, bongo, flute, horn, saxophone, violin \\
\midrule
$\mathcal{T}$ & \textbf{weapons} & arrow, club, gun, missile, spear, axe, dagger, harpoon, pistol, sword, blade, dynamite, hatchet, rifle, tank, bomb, firearm, knife, shotgun, teargas, cannon, grenade, mace, slingshot, whip \\
\midrule
$\mathcal{A}$ & \textbf{pleasant} & caress, freedom, health, love, peace, cheer, friend, heaven, loyal, pleasure, diamond, gentle, honest, lucky, rainbow, diploma, gift, honor, miracle, sunrise, family, happy, laughter, paradise, vacation \\
\midrule
$\mathcal{B}$ & \textbf{unpleasant} & abuse, crash, filth, murder, sickness, accident, death, grief, poison, stink, assault, disaster, hatred, pollute, tragedy, divorce, jail, poverty, ugly, cancer, kill, rotten, vomit, agony, prison \\
\midrule
\end{tabular}
\end{center}
\bigskip

\section{Detailed Experimental Methodology}
Here we detail the experimental methodology used to test our method's accuracy.

\xhdr{\emph{I} - Train a baseline}
We start by training 10 word embeddings using the parameters in Table \ref{tab:setup} above, but using different random seeds.
These embeddings create a baseline for the unperturbed bias $B(w^*)$.

\xhdr{\emph{II} - Approximate the differential bias of each document}
For each WEAT test, we approximate the differential bias of every document in the corpus.
We do so with a combination of Equations (8) and (5) of the main text. 
This step is summarize by Algorithm 1 in the main text.
Note that we make the differential bias approximation for each document several times, using the learned parameters $w^*$, $u^*$, $b^*$ and $c^*$ from the 10 different baseline embeddings in our different approximations.
We then average these approximations for each document, and construct a histogram.

\xhdr{\emph{III} - Construct perturbation sets}
We perturb the corpus by removing sets of documents. We construct three types of perturbation sets: \emph{increase}, \emph{random}, and \emph{decrease}.
The targeted (increase, decrease) perturbation sets are constructed from the documents whose removals were predicted to cause the greatest differential bias (in absolute value), i.e., the documents located in the tails of the histograms. For the Wiki setup we consider the 10, 30, 100, 300, and 1000 most influential documents for each bias, while for the NYT setup we consider the 100, 300, 1000, 3000, and 10,000 most influential. This results in 10 perturbations sets per corpus per bias, for a total of 40.

The random sets are, as their name suggests, drawn uniformly at random from the entire set of documents used in the training corpus.
For the Wiki setup we consider 6 sets of 10, 30, 100, 300, and 1000 documents (30 total). Because training times are much longer, we limit this to 6 sets of 10,000 documents for the NYT setup. Therefore we consider a total of 36 random sets.

\xhdr{\emph{IV} - Approximate the differential bias of each perturbation set}
We then approximate the differential bias of each perturbation set. Note that $\nabla_w L(X_i, w)$ is not linear in $X_i$. Therefore determining the differential bias of a perturbation set does not amount to simply summing the differential bias of each document in the set (although in practice we find it to be close). Here we also make 10 approximations, one with each of the different baseline embeddings.

\xhdr{\emph{V} - Construct ground truth and assess}
Finally, for each perturbation set, we remove the target documents from the corpus, and train 5 new embeddings on this perturbed corpus.
We use the same hyperparameters, again varying only the random seed. 
The bias measured in these embeddings serve as the ground truth for assessment.


\section{Additional experimental results} 
Here we include additional experimental results.
\bigskip

\begin{figure}[h]
    \centering
    \includegraphics[width=65mm]{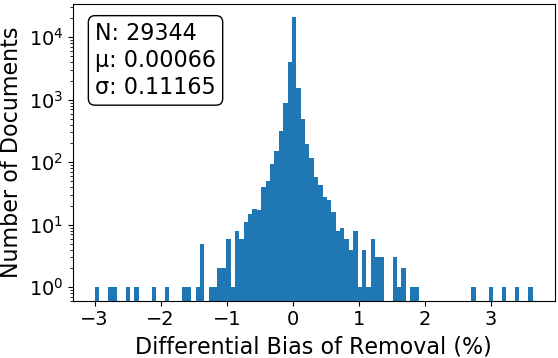}
    \includegraphics[width=65mm]{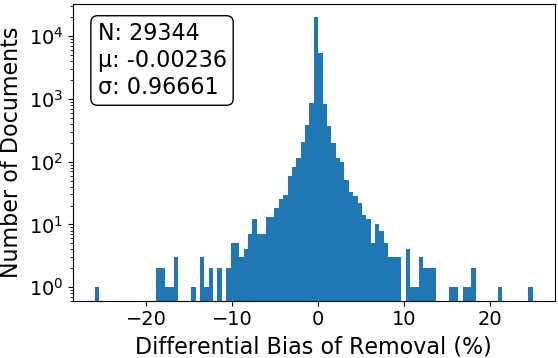}
    \includegraphics[width=65mm]{figures/histogram_NYT_1.png}
    \includegraphics[width=65mm]{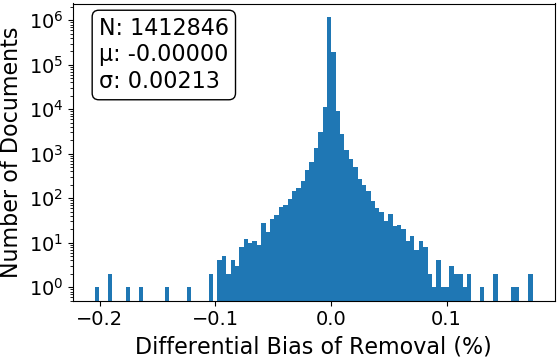}
    \caption{Histogram of the approximated differential bias of removal for every document in our Wiki setup (top) and NYT setup (bottom), considering WEAT1 (left) and WEAT2 (right), measured in percent change from the corresponding mean baseline bias.}
\end{figure}

\begin{figure}
  \centering
  \includegraphics[width=65mm]{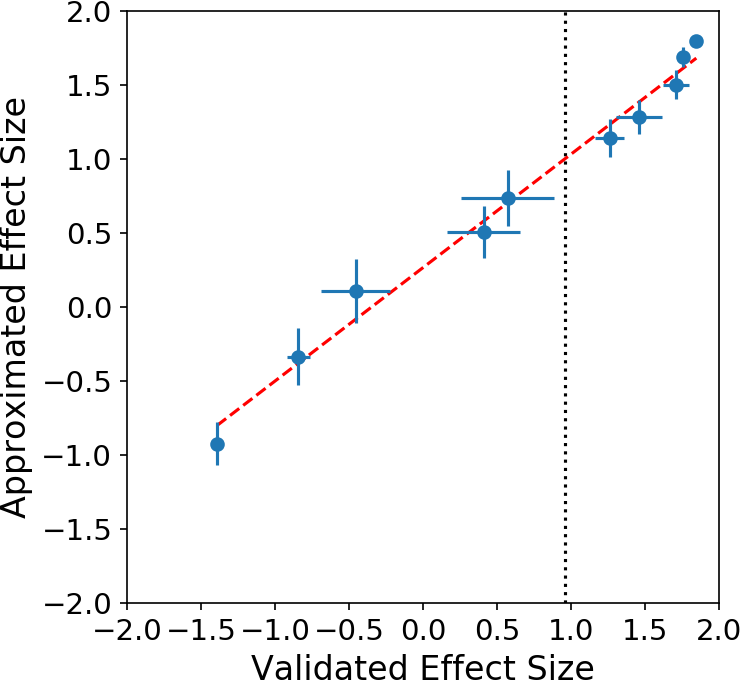}
  \includegraphics[width=65mm]{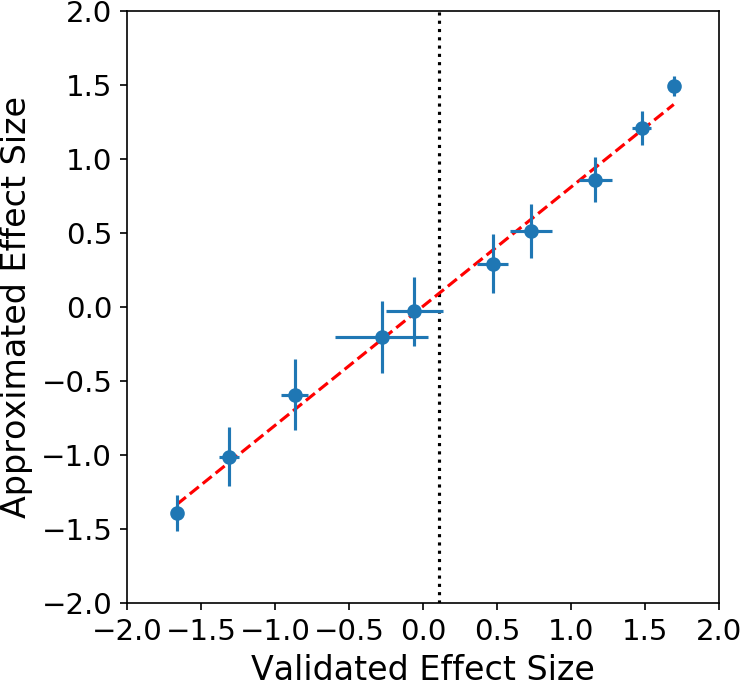}
  \includegraphics[width=65mm]{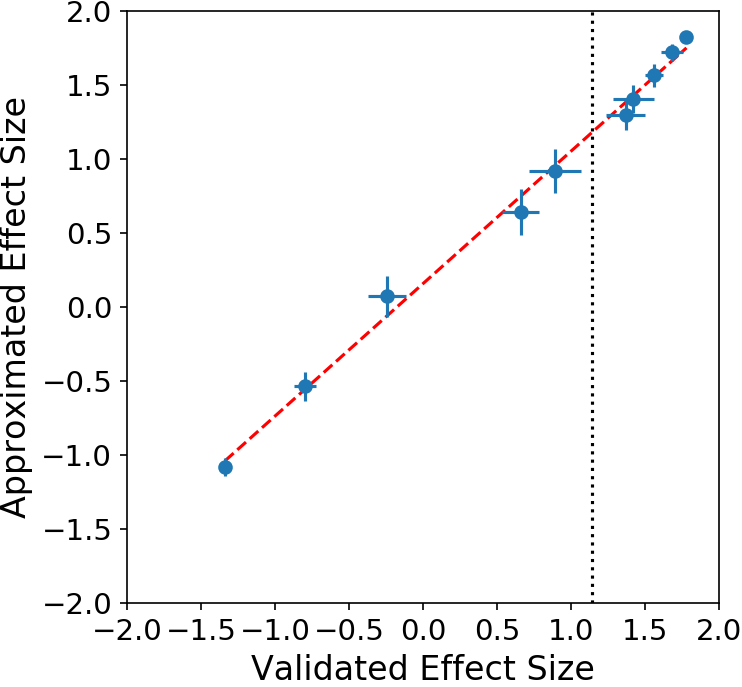}
  \includegraphics[width=65mm]{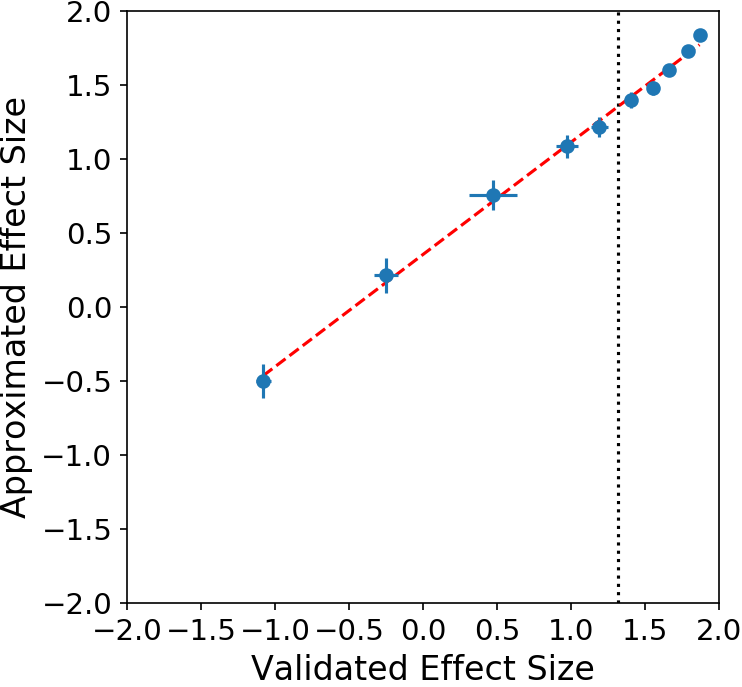}
  \caption{Approximated vs. ground truth WEAT bias effect size due to the removal of each (non-random) perturbation set in Wiki setup (top) and NYT setup (bottom), considering WEAT1 (left) and WEAT2 (right); points plot the means; error bars depict one standard deviation; dashed line shows least squares; the baseline means are shown with vertical dotted lines; correlations in Table \ref{tab:correlation}. }
\end{figure}

\begin{figure}
  \centering
   \includegraphics[width=65mm]{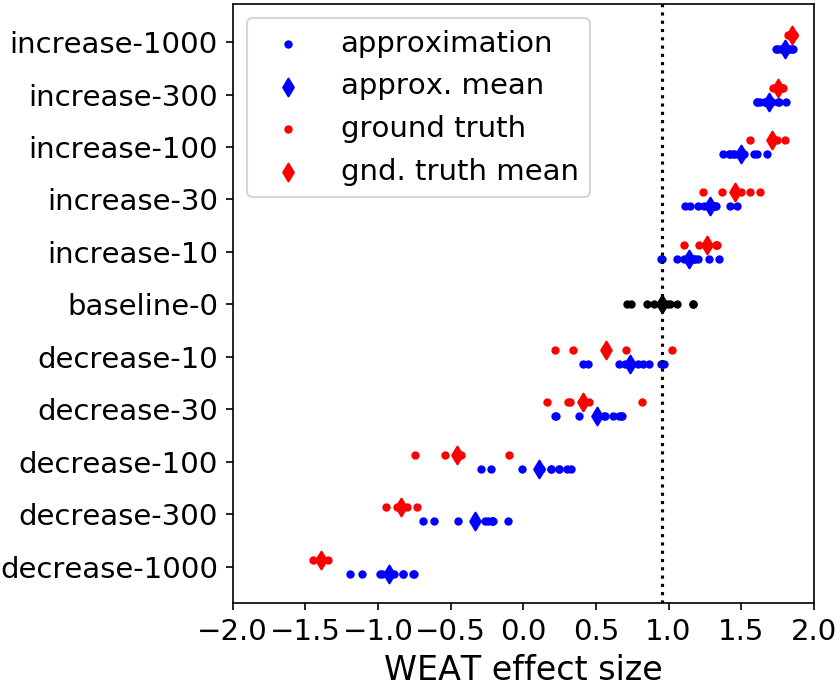}
  \includegraphics[width=65mm]{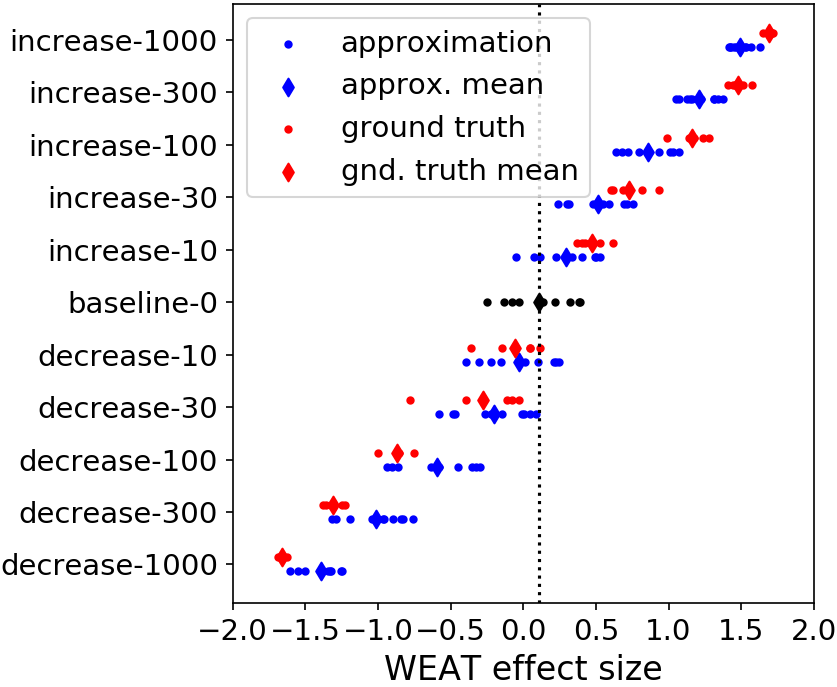}
  \includegraphics[width=65mm]{figures/targeted_NYT_1.png}
  \includegraphics[width=65mm]{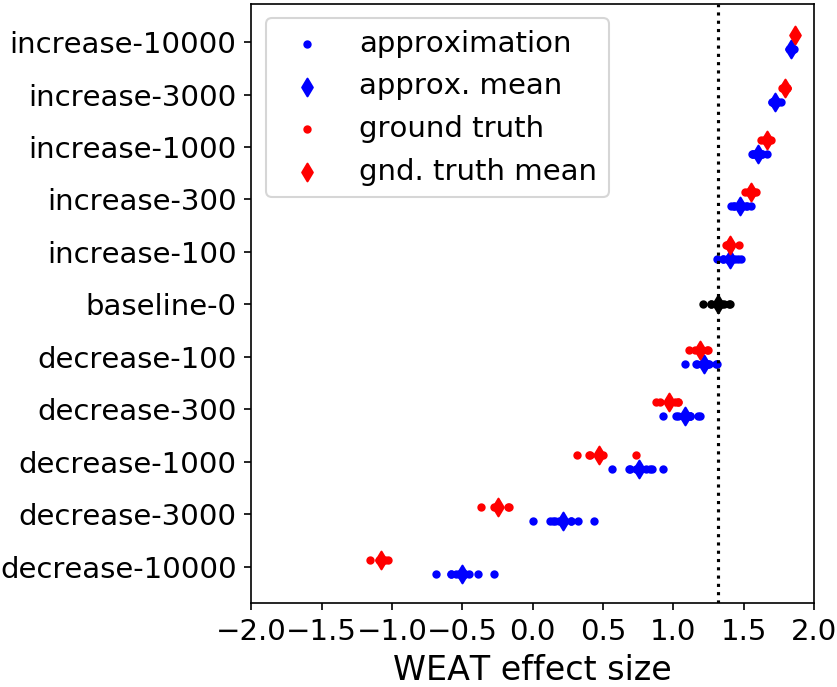}
  \caption{Approximated and ground truth differential bias of removal for every (non-random) perturbation set in Wiki setup (top) and NYT setup (bottom), considering WEAT1 (left) and WEAT2 (right); the baseline means are shown with vertical dotted lines}
\end{figure}

\begin{table}[h]
\caption{Correlation of Approximated and Validated Mean Biases}
\label{tab:correlation}
\centering
\begin{tabular}{ l  c  c }
\toprule
     & WEAT1        & WEAT2 \\ \midrule
Wiki & $r^2$: 0.986 & $r^2$: 0.993\\
NYT  & $r^2$: 0.995 & $r^2$: 0.997\\
\end{tabular}
\end{table}

\begin{figure}
  \centering
  \includegraphics[width=65mm]{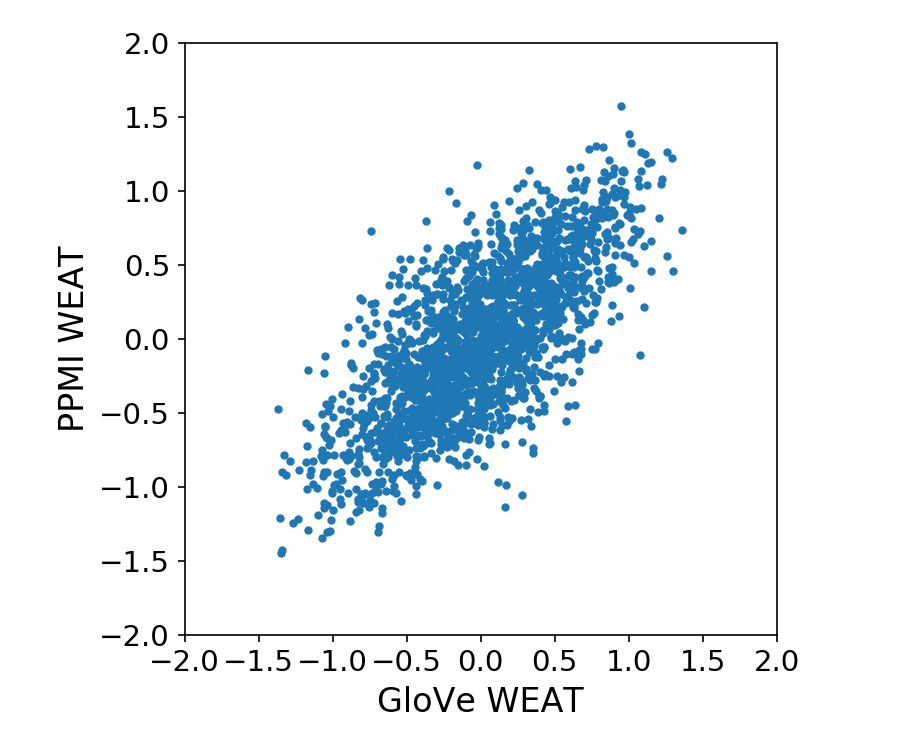}
  \includegraphics[width=65mm]{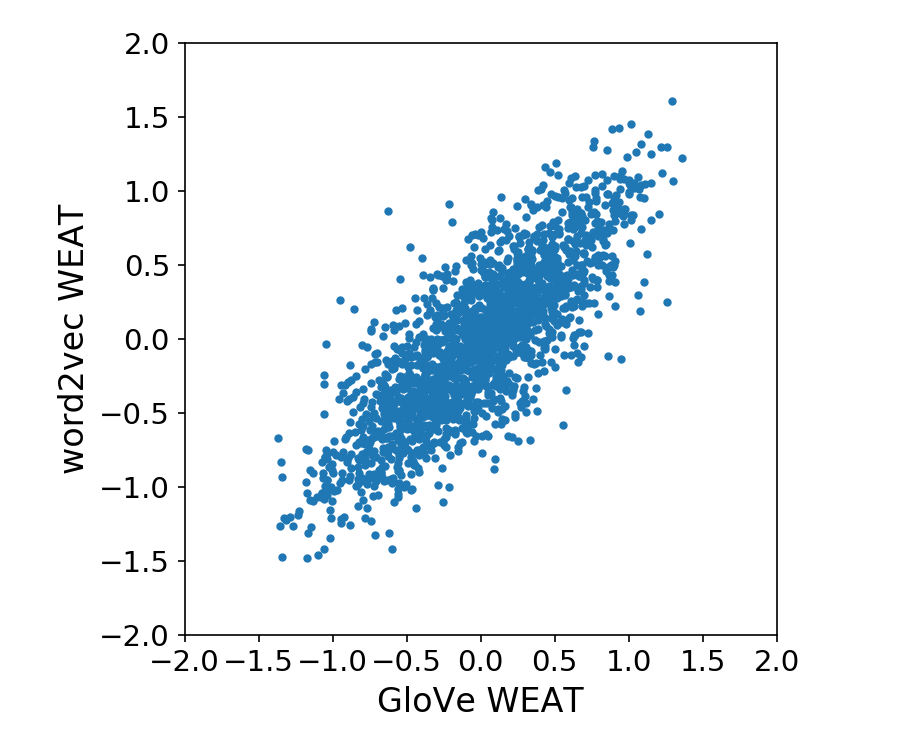}
  \caption{The correlation of the WEAT as measured in our NYT GloVe embeddings versus the corpus' PPMI representation in 2000 randomly generated word sets, $r^2=0.725$ (left); versus when measured in word2vec embeddings with comparable hyper-parameters, $r^2=0.803$ (right).}
\end{figure}

\begin{table}
\caption{A comparison of the effect of removing the most impactful documents as identified by a PPMI baseline technique versus when identified by our method (Wiki setup, mean of WEAT1 in 10 retrained GloVe embeddings). }
\label{tab:ppmi_compare}
\centering
\begin{tabular}{ l c c  c }
\toprule
\multicolumn{2}{c}{Document Set}    & \multicolumn{2}{c}{$\Delta B$ when Identified by} \\
objective        & num. docs.            & baseline  & our method \\ 
\midrule
correct     & 300                   & -67\% & -187\% \\
correct     & 100                    & -50\% & -147\% \\
correct     & 30                    & -23\% & -57\% \\
correct     & 10                    & -4\% & -40\% \\
aggravate   & 10                    & -0.5\%  & 32\% \\
aggravate   & 30                    & 20\%  & 53\% \\
aggravate   & 100                    & 15\%  & 79\% \\
aggravate   & 300                   & 47\% & 84\% \\
\end{tabular}
\end{table}


\newpage

\section{Influential Documents - NYT WEAT 1}
The below documents were identified to be the 50 most WEAT1 bias influencing documents in our NYT setup. We list the article titles. Publication dates range from January 1, 1987 to June 19, 2007. Most can be found through \textit{https://www.nytimes.com/search}. A subscription may be required for access.

\vfill

\begin{center}

\begin{tabular}{ l p{8cm} }
\toprule
$\Delta_{\text{doc}} B$ & \textbf{Bias Decreasing} \\
\midrule
-0.52	& Hormone Therapy Study Finds Risk for Some \\
-0.50	& For Women in Astronomy, a Glass Ceiling in the Sky \\
-0.49	& Sorting Through the Confusion Over Estrogen \\
-0.36	& Young Astronomers Scan Night Sky and Help Wanted Ads \\
-0.33	& Campus Where Stars Are a Major \\
-0.33	& A New Look At Estrogen And Stroke \\
-0.31	& Scenes From a Space Thriller \\
-0.30	& The Cosmos Gets Another Set of Eyes \\
-0.29	& The Stars Can't Help It \\
-0.28	& Making Science Fact, Now Chronicling Science Fiction \\
-0.27	& Estrogen Heart Study Proves Discouraging \\
-0.26	& EINSTEIN LETTERS TELL OF ANGUISHED LOVE AFFAIR \\
-0.25	& Divorcing Astronomy \\
-0.24	& Astronomers Open New Search for Alien Life \\
-0.23	& AT WORK WITH: Susie Cox; Even Stars Need a Map To the Galaxy \\
-0.22	& CAMPUS LIFE: Minnesota; Astronomer Spots Clue To Future of Universe \\
-0.21	& Clothes That Are Colorful and TV's That Are Thin Make Many Lists \\
-0.20	& We Are the Fourth World \\
-0.20	& Hitched to a Star, With a Go-To Gadget \\
-0.19	& Material World \\
-0.19	& Shuttle's Stargazing Disappoints Astronomers \\
-0.19	& 2 Equity Firms Set to Acquire Neiman Marcus \\
-0.18	& Volunteer's Chain Letter Embarrasses a Hospital \\
-0.18	& What Doctors Don't Know (Almost Everything) \\
-0.18	& Astronomers Edging Closer To Gaining Black Hole Image \\
\end{tabular}

\vfill

\pagebreak

\hspace{0pt}
\vfill

\begin{tabular}{ l p{8cm} }
\toprule
$\Delta_{\text{doc}} B$ & \textbf{Bias Increasing} \\
\midrule
0.38	& Kaj Aage Strand, 93, Astronomer At the U.S. Naval Observatory \\
0.32	& Gunman in Iowa Wrote of Plans In Five Letters \\
0.29	& ENGINEER WARNED ABOUT DIRE IMPACT OF LIFTOFF DAMAGE \\
0.29	& Fred Gillett, 64; Studied Infrared Astronomy \\
0.27	& Robert Harrington, 50, Astronomer in Capital \\
0.27	& For Voyager 2's 'Family' of 17 Years, It's the Last of the First Encounters \\
0.26	& Despite the Light, Astronomers Survive \\
0.25	& LONG ISLAND GUIDE \\
0.25	& THE GUIDE \\
0.24	& Telescope Will Offer X-Ray View Of Cosmos \\
0.23	& Astronomers Debate Conflicting Answers for the Age of the Universe \\
0.23	& The Wild Country of Anza Borrego \\
0.21	& What Time Is It in the Transept? \\
0.21	& Jan H. Oort, Dutch Astronomer In Forefront of Field, Dies at 92 \\
0.21	& Logging On to the Stars \\
0.20	& The Sky, Up Close and Digital \\
0.20	& Q\&A \\
0.19	& Getting Attention With Texas Excess \\
0.19	& Emily's College \\
0.19	& 60 New Members Elected to Academy of Sciences \\
0.18	& Theoretical Physics, in Video: A Thrill Ride to 'the Other Side of Infinity' \\
0.18	& Charles A. Federer Jr., Stargazer-Editor, 90 \\
0.18	& Some Web sites are taking their brands from the Internet into some very offline spheres. \\
0.18	& A Wealth of Cultural Nuggets Waiting to Be Mined \\
0.18	& Can a Robot Save Hubble? More Scientists Think So \\

\end{tabular}
\vfill

\end{center}

\pagebreak
\section{Influence of Mulitple Perturbations}
Here we show how we can extend the influence function equations presented by Koh \& Liang (2017) to address the case of multiple training point perturbations. We do not intend this to be a rigorous mathematical proof, but rather to provide insight into the logical steps we followed. 

First we summarize the derivation in the case of a single train point perturbation. Let  $R(z, \theta)$ be a convex scalar loss function for a learning task, with optimal model parameters $\theta^*$ of the form in Equation \ref{eq:model_2} below, where $\{z_1, ... , z_n\}$ are the training data points and $L(z_i, \theta)$ is the point-wise loss.
\begin{align}\label{eq:model_2}
R(z, \theta) = \frac{1}{n} \sum_{i=1}^n L(z_i, \theta) && \theta^* = \argmin_{\theta} R(z, \theta)
\end{align}

We would like to determine how the optimal parameters $\theta^*$ would change if we perturbed the $k^{th}$ point in the training set; i.e., $z_k \rightarrow \tilde{z}_k$. The optimal parameters under perturbation can be written as:
\begin{align}\label{eq:theta_eps}
\tilde{\theta}(\varepsilon) = \argmin_{\theta} \bigg\{ R(z, \theta) + \varepsilon L(\tilde{z}_k, \theta) - \varepsilon L(z_k, \theta) \bigg\}
\end{align}
where we seek $\tilde{\theta}\rvert_{\varepsilon=\frac{1}{n}}$, noting that $\tilde{\theta}\rvert_{\varepsilon=0} = \theta^*$. Since $\tilde{\theta}$ minimizes Equation \ref{eq:theta_eps}, we must have
\begin{align*}
0 = \nabla_{\theta} R(z, \tilde{\theta}) + \varepsilon \nabla_{\theta} L(\tilde{z}_k, \tilde{\theta}) - \varepsilon \nabla_{\theta} L(z_k, \tilde{\theta})
\end{align*}
for which we can compute the first order Taylor series expansion (with respect to $\theta$) around $\theta^*$. This gives:
\begin{align*}
0 \approx& \nabla_{\theta} R(z, \theta^*)+  \varepsilon \nabla_{\theta} L(\tilde{z}_k, \theta^*) -  \varepsilon \nabla_{\theta} L(z_k, \theta^*) \\
&+ \big[\nabla_{\theta}^2 R(z, \theta^*) + \varepsilon \nabla_{\theta}^2 L(\tilde{z}_k, \theta^*) - \varepsilon \nabla_{\theta}^2 L(z_k, \theta^*)\big](\tilde{\theta} - \theta^*)
\end{align*}

Noting $\nabla_{\theta} R(z, \theta^*) = 0$, then keeping only $\BigO{\varepsilon}$ terms, solving for $\tilde{\theta}$, and evaluating at $\varepsilon = \frac{1}{n}$ we obtain:
\begin{align}\label{eq:influence_2}
\tilde{\theta} - \theta^* \approx
\left( \frac{-1}{n} \right) \, H_{\theta^*} ^{-1} \left[ \nabla_{\theta} L(\tilde{z}_k, \theta^*) - \nabla_{\theta} L(z_k, \theta^*)\right]
\end{align}
where $H_{\theta^*} = \frac{1}{n}\sum_{i=1}^n \nabla_{\theta}^2 L(z_i,\theta^*)$ is the Hessian of the total loss.

Now, we address the more general case where several training points are perturbed. 
This corresponds to replacing the expression $\varepsilon L(\tilde{z}_k, \theta) - \varepsilon L(z_k, \theta)$ in Equation \eqref{eq:theta_eps} with $\sum_{k\in\delta} \big( \varepsilon L(\tilde{z}_k, \theta) - \varepsilon L(z_k, \theta) \big)$, where $\delta$ is the set of indices of perturbed points. 
Because of the linearity of the gradient operator, we can readily carry this substitution through the subsequent equations, resulting in:
\begin{align}\label{eq:influence_sum}
\tilde{\theta} - \theta^* \approx
\left( \frac{-1}{n} \right) \, H_{\theta^*} ^{-1} \sum_{k\in\delta}\left[\nabla_{\theta} L(\tilde{z}_k, \theta^*) - \nabla_{\theta} L(z_k, \theta^*)\right]
\end{align}
where we assume $|\delta| \ll n$.

\end{document}